\documentclass{article}

\PassOptionsToPackage{numbers, compress}{natbib}
\usepackage[preprint]{neurips_2023}

\usepackage[utf8]{inputenc} %
\usepackage[T1]{fontenc}    %
\usepackage{hyperref}       %
\usepackage{url}            %
\usepackage{booktabs}       %
\usepackage{amsfonts}       %
\usepackage{nicefrac}       %
\usepackage{microtype}      %
\usepackage{xcolor}         %

\usepackage{graphicx}
\usepackage{booktabs}
\usepackage{multirow}
\usepackage{xcolor}
\usepackage{xspace}
\usepackage{subfig}

\newif\ifcomments
\commentstrue
\ifcomments
\usepackage[textsize=small,color=lightgray]{todonotes}
\else
\usepackage[disable]{todonotes}
\fi

\newcommand{\fref}[1]{Figure~\ref{#1}}
\newcommand{\tref}[1]{Table~\ref{#1}}
\newcommand{\secref}[1]{Section~\ref{#1}}
\newcommand{\eqref}[1]{Equation~\ref{#1}}

\newcounter{tacounter} %
\newcommand{\takeaway}[1]{\textbf{Takeaway \stepcounter{tacounter}\thetacounter:} \textit{#1}}

\newcommand{\dirich}[1]{$\mathrm{Dir}(#1)$}

\newcommand{\newtext}[1]{{\color{blue} #1}}
\newcommand{\drop}[1]{}

\title{Not All Federated Learning Algorithms Are Created Equal: A Performance Evaluation Study}

\author{%
Gustav A. Baumgart$^\dag$ \, Jaemin Shin$^\ddag$ \, Ali Payani$^*$ \, Myungjin Lee$^*$ \, Ramana Rao Kompella$^*$\\
  $^\dag$University of Minnesota \quad $^\ddag$KAIST \quad $^*$Cisco Research\\
  \texttt{baumg260@umn.edu}\\
  \texttt{jaemin.shin@kaist.ac.kr}\\
  \texttt{\{apayani, myungjle, rkompell\}@cisco.com}
}

\begin{document}

\maketitle

\begin{abstract}
\label{sec:abstract}

Federated Learning (FL) emerged as a practical approach to training a model from decentralized data. The proliferation of FL led to the development of numerous FL algorithms and mechanisms. Many prior efforts have given their primary focus on accuracy of those approaches, but there exists little understanding of other aspects such as computational overheads, performance and training stability, etc. To bridge this gap, we conduct extensive performance evaluation on several canonical FL algorithms (FedAvg, FedProx, FedYogi, FedAdam, SCAFFOLD, and FedDyn) by leveraging an open-source federated learning framework called Flame. Our comprehensive measurement study reveals that no single algorithm works best across different performance metrics. A few key observations are: (1) While some state-of-the-art algorithms achieve higher accuracy than others, they incur either higher computation overheads (FedDyn) or communication overheads (SCAFFOLD). (2) Recent algorithms present smaller standard deviation in accuracy across clients than FedAvg, indicating that the advanced algorithms' performances are stable. (3) However, algorithms such as FedDyn and SCAFFOLD are more prone to catastrophic failures without the support of additional techniques such as gradient clipping. We hope that our empirical study can help the community to build best practices in evaluating FL algorithms.

\end{abstract}

\section{Introduction}
\label{sec:introduction}

Machine learning (ML) is fueling modern applications, and high-performance ML models are becoming key to business success. However, ML training often requires a large volume of high quality datasets and centralized learning can violate data privacy. Federated learning (FL) emerged as a practical approach that can train machine learning models from decentralized data while preserving privacy.
Its potential drew tremendous attention from the research community, which resulted in the invention of numerous FL algorithms and mechanisms~\cite{fl_survey_tkde, fedavg, feddyn, fedprox, fedopt, scaffold, feddc, oort, fedbuff, fedbalancer, mohawk}.

Many prior works focus on performance metrics such as accuracy, theoretical convergence rate, fairness, and communication efficiency.
For instance, some FL algorithm studies~\cite{feddyn, feddc, fedprox, fedavg} focus on accuracy while others~\cite{abay2020mitigating, chu2021fedfair, ezzeldin2021fairfed, fedgft} evaluate group fairness. To discuss communication efficiency and speedup of algorithms, most studies measure the number of rounds to reach a certain accuracy as their key metric~\cite{scaffold, feddyn, feddc, fedopt}.

However, such a performance evaluation of previous approaches provide little insight into the performance of FL algorithms under realistic scenarios, without considering actual wall-clock time or computational
overheads. For example, previous studies reporting the number of rounds to convergence neglect computation and communication overheads per round. While there were few studies~\cite{oort, fedbalancer} that measured the wall-clock time and resource usage of FL algorithms (time-to-accuracy and resource-to-accuracy), they were conducted at simulation-based environments without actual model weights communication in FL. Thus, previous analysis significantly limits understanding of the FL approaches in a real-world setting and makes it difficult for the FL practitioners to choose an appropriate approach under different constraints.

In this paper, we argue for the need for a holistic evaluation of FL approaches, which is imperative to assist FL practitioners better. While there may be many aspects that the practitioners want to understand for a given approach, we primarily consider four aspects: (i) computation costs, (ii) communication overheads, (iii) performance stability across clients, and (iv) training instability. 
First, computation costs can influence time-to-accuracy depending on the algorithm's complexity.
Second, in addition to model weights, some algorithms ask for the exchange of extra information between clients and a server. This incurs different degrees of communication overheads and could bottleneck FL, thereby affecting time-to-accuracy. Third, the global model produced by an FL job should maintain a similar performance across different clients. We call this property performance stability across clients.
There is a lack of research on the performance stability.
The instability can potentially harm an application's credibility as some users may experience poorer performance compared to others. Finally, instability can occur during training due to various reasons in real FL deployments, which we call training instability. As a means to explore the training instability of the algorithms, we choose gradient clipping~\cite{goodfellow2016deep}
The technique and its variants
have been adopted in FL settings~\cite{feddyn, cohort}.
The main idea behind the technique is to prevent overflow for the model weights by controlling large client updates (from a batch in local training or from a complete round).
However, it is unknown how the absence of gradient clipping would affect these algorithms.

To conduct a comprehensive evaluation study on FL approaches, we consider various factors: algorithm, accelerator\newtext{,} and model architecture. First, we focus on several canonical algorithms (FedAvg~\cite{fedavg}, server-side optimizer~\cite{fedopt} (e.g., FedAdam and FedYogi), SCAFFOLD~\cite{scaffold}, FedProx~\cite{fedprox}, and FedDyn~\cite{feddyn}). Second, we rely on different hardware settings (e.g., CPU and NVIDIA GPUs such as A100, V100 and T4)
to check how the algorithms behave under different resource constraints. Third, we employ different model architectures such as CNN, ResNet, and LSTM. We use CNN and ResNet for training with the CIFAR-10 dataset and LSTM for Shakespeare dataset. We leverage 
Flame\footnote{https://github.com/cisco-open/flame}~\cite{flame2023},
an open-source federated learning framework, to run experiments by combining those factors. Its well-defined programming interface allows us to implement various algorithms consistently. In turn, the evaluation of the algorithms under a \textit{single} framework renders our results easily reproducible.

Our comprehensive performance evaluation study makes the following key contributions:
\begin{itemize}

\item Performance metrics based on round (such as accuracy-to-round, the accuracy achieved after a certain number of rounds) 
should be interpreted carefully.
Such metrics suggest that the smaller number of rounds is indicative of high performance. This ignores the amount of computations (hence, the actual walk-clock time) needed for training.
Our experiments show that FedDyn achieves higher accuracy for 100 rounds of training than other algorithms but takes approximately 1.58$\times$ as long compared to FedAvg (see \secref{subsec:general_performance} and \fref{fig:tst_acc_cifar10_dir0.3}).

\item FL algorithms exhibit different degrees of computation overheads depending on hardware and model architecture (\secref{s:runtime}). In general, the state-of-the-art algorithms such as SCAFFOLD, FedProx, and FedDyn tend to have significant runtime increases over FedAvg (\fref{fig:cifar10_runtimes}) when CNN and ResNet are employed. %
In particular, when resources are constrained, these algorithms incur much higher runtime (e.g., FedDyn has a 252.90\% runtime increase over FedAvg as shown in \fref{fig:cifar10_runtimes_cpu}).
In the case of LSTM, however, \fref{fig:shakes_runtimes} shows that the less the resources are, the less runtime gap is between algorithms. With an A100 GPU, FedDyn's runtime is about 40\% higher than that of FedAvg (\fref{fig:shakes_runtimes_a100}). In contrast, under a CPU, the gap is less than 5\% across the algorithms; even FedDyn is 10\% faster than that of FedAvg (\fref{fig:shakes_runtimes_cpu}).
This indicates FL algorithms runtime is influenced by model architectures.

\item Algorithms exhibit different levels of performance stability 
across clients (\secref{s:stability}). Our experiment results allow us to make several observations. First, FedDyn achieves the best performance stability among all the algorithms. In other words, it delivers more consistent local test accuracy results across clients.
Second, SCAFFOLD is more vulnerable to class imbalances seen in non-IID data. It tends to show higher standard deviation in accuracy than FedAvg. Third, server-side optimization algorithms such as FedAdam and FedYogi can be alternatives to client-side optimization algorithms as they are lightweight and often obtain better performance stability than SCAFFOLD and FedProx.
The violin plots for local test accuracies in \fref{fig:cifar10_psac} and \fref{fig:shakespeare_psac} support these observations.

\item Client-side optimization algorithms are more vulnerable to catastrophic failures.
As a means for testing, we conduct experiments by disabling gradient clipping.
The experiment results in \tref{tbl:grad_clip_two} show that as the dataset distribution becomes heterogeneous, SCAFFOLD and FedDyn experience more frequent failures while other algorithms experience no failure.
More specifically, we observe that only FedDyn and SCAFFOLD face failure rates at or above 60\%. Notably, a smaller learning rate helps these failure rates drop to 0\%. Therefore, it is essential to use gradient clipping for algorithms such as FedDyn and SCAFFOLD.

\item To conduct our performance evaluation study, we implement many of the algorithms used in this paper within Flame~\cite{flame2023}. We make these algorithms available in the Flame's open-source repository, which can help reproduce the evaluation results in this work reliably.

\end{itemize}

\section{Related Work}
\label{sec:related}

\paragraph{Federated Learning Approaches.} 
Upon the introduction of Federated Learning (FL)~\cite{konevcny2016federated}, numerous approaches have been subsequently proposed to optimize FL at various aspects~\cite{feddyn, feddc, scaffold, fedprox, qfedavg, fedavg, fedbuff, fedopt, fednova}. Among these proposals, we select six widely adopted approaches for extensive performance evaluation as follows: (1) \textit{FedAvg}~\cite{fedavg}, the most commonly used FL approach~\cite{jiang-sensors2020}, has been proposed to optimize FL communication efficiency by training multiple local epochs on clients instead of one. (2) Li et al.~\cite{fedprox} proposed \textit{FedProx}, which adds a client-side regularization for robust training on clients with heterogeneous non-IID data distributions. (3) \textit{FedAdam} and (4) \textit{FedYogi} were proposed by Reddi et al.~\cite{fedopt}. These algorithms provide adaptivity at server-side optimization to improve FL convergence. (5) \textit{SCAFFOLD}~\cite{scaffold} was also proposed to achieve improved convergence rates by reducing the variance among clients with heterogeneous data, using control variates to correct local model updates. (6) \textit{FedDyn}~\cite{feddyn} focuses on aligning local and global objectives in FL through a client-side regularizer, mitigating the impact of data heterogeneity and improving the convergence rate. In this paper, we report the extensive experimental comparison of these approaches on model performance and system-level metrics.

\paragraph{Practical Evaluation of Federated Learning Approaches.} Several prior studies have approached to evaluate the performance of the FL approaches as follows: Reddi et al.~\cite{fedopt} compared the model performance and hyperparameter sensitivity across the aforementioned approaches, except for FedDyn, on four datasets. Gao et al.~\cite{feddc} measured the model performance and number of rounds for convergence of the approaches on six non-iid datasets. Charles et al.~\cite{cohort} investigated the effects of the number of clients sampled per round in FL, examining aspects such as model failures, convergence, and fairness. However, these results predominantly focus on model performance through simulation-based experiments without actual model weights communication or computation time measurement. This approach overlooks crucial practical considerations, such as the evaluation of system-level metrics like training and communication overhead. There exist studies~\cite{fedscale, oort, fedbalancer, 10.1145/3442381.3449851} that approached to simulate train and communication latency of clients by leveraging data from 136k user mobile devices. However, these studies rest on the assumption that client time distributions remain constant, which may not hold in dynamic real-world environments where train and computation latency can fluctuate significantly. Moreover, their experiments only partially incorporated the FL approaches that we consider in this work (FedAvg, FedProx). %

Recognizing these gaps, our work aims to present a comprehensive measurement study of the six widely adopted FL approaches. We focus not just on traditional metrics on model performance but also on system-level factors, employing a testbed that facilitates actual model training and weight communication in FL. This approach is designed to provide a more realistic and holistic assessment of FL approaches, addressing practical aspects that previous studies have overlooked.

\section{Preliminaries}
\label{sec:preliminaries}

In this section, we first discuss the algorithms we use in our evaluation. We choose six different algorithms: FedAvg, FedYogi, FedAdam, FedProx, SCAFFOLD and FedDyn.
We discuss the loss functions of these algorithms in \secref{subsec:algo}. We exclude algorithms that are particularly tailored for certain metrics (e.g., fairness); we leave the evaluation on those algorithms as future work.
Next, in \secref{subsec:flame}, we briefly describe a federated learning framework called Flame we rely on to facilitate realistic experiments. Flame isolates individual components (client and server) into separate processes and lets them run in parallel, which can demonstrate concurrent execution of clients' training that takes place in a real-world setting.

\subsection{FL Algorithms}
\label{subsec:algo}

\textbf{FedAvg~\cite{fedavg}.} FedAvg is the first aggregation algorithm that employs
local stochastic gradient descent (SGD) in a federated learning setting.
This approach is susceptible to non-IID datasets, suffering from an inferior convergence rate
in a non-IID setting to its convergence rate in an IID setting.
In FedAvg's loss function, there is no extra correction (regularization) term. We let $L(\theta)$ denote FedAvg's loss function, where $\theta$ represents the model weights.

\textbf{FedYogi~\cite{fedopt}} and \textbf{FedAdam~\cite{fedopt}.} They attempt to address the issue of heterogeneous data.
They additionally introduce server-side aggregation formulae that exploit second-degree approximation during the aggregation step at the server.
At the client side, the training process is the same as FedAvg, meaning they both have the same loss function $L(\theta)$.
They neither require any additional communication nor does a server-side state need to be maintained throughout the training process at the client side (besides the global model).

\textbf{FedProx~\cite{fedprox}.} This algorithm adds a regularization term to the loss function at the client side, which would essentially keep the updates closer to global model at that round. The right-hand side term in \eqref{fedprox_loss} discourages large client-side updates by minimizing over the squared norm of the distance between the local model weights and the current global model weights.
Here, $\theta^t$ represents the current global model's weights.
\begin{equation}\label{fedprox_loss}
L'(\theta) = L(\theta) + \frac{\mu}{2}||\theta - \theta^t||^2
\end{equation}

\textbf{SCAFFOLD~\cite{scaffold}.} This algorithm converges faster than FedAvg under heterogeneous settings~\cite{scaffold}.
Although the modifications to client-side training are not as significant as FedProx, this method requires two times as much communication volume per round because communication includes the control variates, which have the same size as the model weights.
The server keeps track of a general gradient direction for all clients $c$, and the client keeps track of its own direction $c_i$.
That way, the term $c-c_i$ is used to correct local training step in order to account for other client's distributions.
Instead of changing it by adding it to the gradient step, we factor this in by modifying the loss function in Equation \ref{scaffold_loss}, which delivers an equivalent result during training.
The operator $\odot$ represents a dot product.
\begin{equation}\label{scaffold_loss}
L'(\theta) = L(\theta) + (c - c_i ) \odot \theta
\end{equation}

\textbf{FedDyn~\cite{feddyn}.} This algorithm is closely related to SCAFFOLD. However, they are different in that FedDyn does not incur the two-fold increase in SCAFFOLD's communication volume. Further, while SCAFFOLD requires additional hyperparameter tuning, the exact minimization technique in FedDyn demands less hyperarameter tuning~\cite{feddyn} .
However, the FedDyn client-side regularization term is more computationally expensive than SCAFFOLD's.
While SCAFFOLD modifies the gradient at each step by a fixed amount (which is equivalent to adding a term linear in $w$ to the loss function), FedDyn adds both linear and quadratic terms.
As shown in \eqref{feddyn_loss}, FedDyn contains two terms; one is the term in FedProx and the other is a linear term similar to what is in SCAFFOLD.
$\nabla L(\theta^{t-1}_k)$ is the local gradient, which is updated last for each client $k$ at the end of training from the last round ($t-1$).
\begin{equation}\label{feddyn_loss}
L'(\theta) = L(\theta) + \nabla L(\theta^{t-1}_k) \odot \theta + \frac{\alpha}{2}||\theta - \theta^t||^2
\end{equation}

\subsection{Flame}
\label{subsec:flame}

We conduct performance evaluation study on FL algorithms and mechanisms in realistic settings to better understand various aspects of the algorithms such as computational overheads, communication costs, performance stability, time to convergence, etc. To  achieve this goal, we need a federated learning framework that facilitates easy developments of algorithms and mechanisms and allows realistic experiments in a real-world testbed.
We choose Flame~\cite{flame2023} as it satisfies our needs. First, its modular design makes it highly extensible and easy to use.
It also lets each component in FL (such as server and client) run as separate process with real communication protocols, thereby enabling realistic experiments. In the remainder of this section, we briefly describe Flame.

Flame's extensibility comes from its abstraction on federated learning components. Flame's abstraction consists of two elements: role and channel. In the abstraction, an FL component is defined as role and communication between components is expressed as channel. For instance, a conventional FL setting has server and client as roles and the connection between the two represents a channel, which means that communication between the two is allowed. In order to enable new mechanisms such as hierarchical FL~\cite{hfl}, defining a new role such as intermediate server and implementing its logic is sufficient while keeping other roles and their implementation intact.
Different FL algorithms may need different protocols to exchange extra information in addition to model weights. Deviations from the basic FL algorithm, FedAvg, can be easily accomplished by overriding existing roles and defining new message types.
We do implement several of the federated learning algorithms (FedYogi, FedAdam, SCAFFOLD, FedProx, and FedDyn) into Flame in a cohesive fashion. %

\section{Evaluation}
\label{sec:evaluation}

We conduct a performance evaluation in terms of various metrics such as accuracy, algorithm and communication overheads, performance stability across clients, and training instability. %
We make use of Flame in realistic settings to accurately reflect on the influences of real systems in running federated learning algorithms.

\begin{figure}[t]
\centering
\subfloat[Test accuracy vs. round]{\includegraphics[width=0.45\columnwidth]{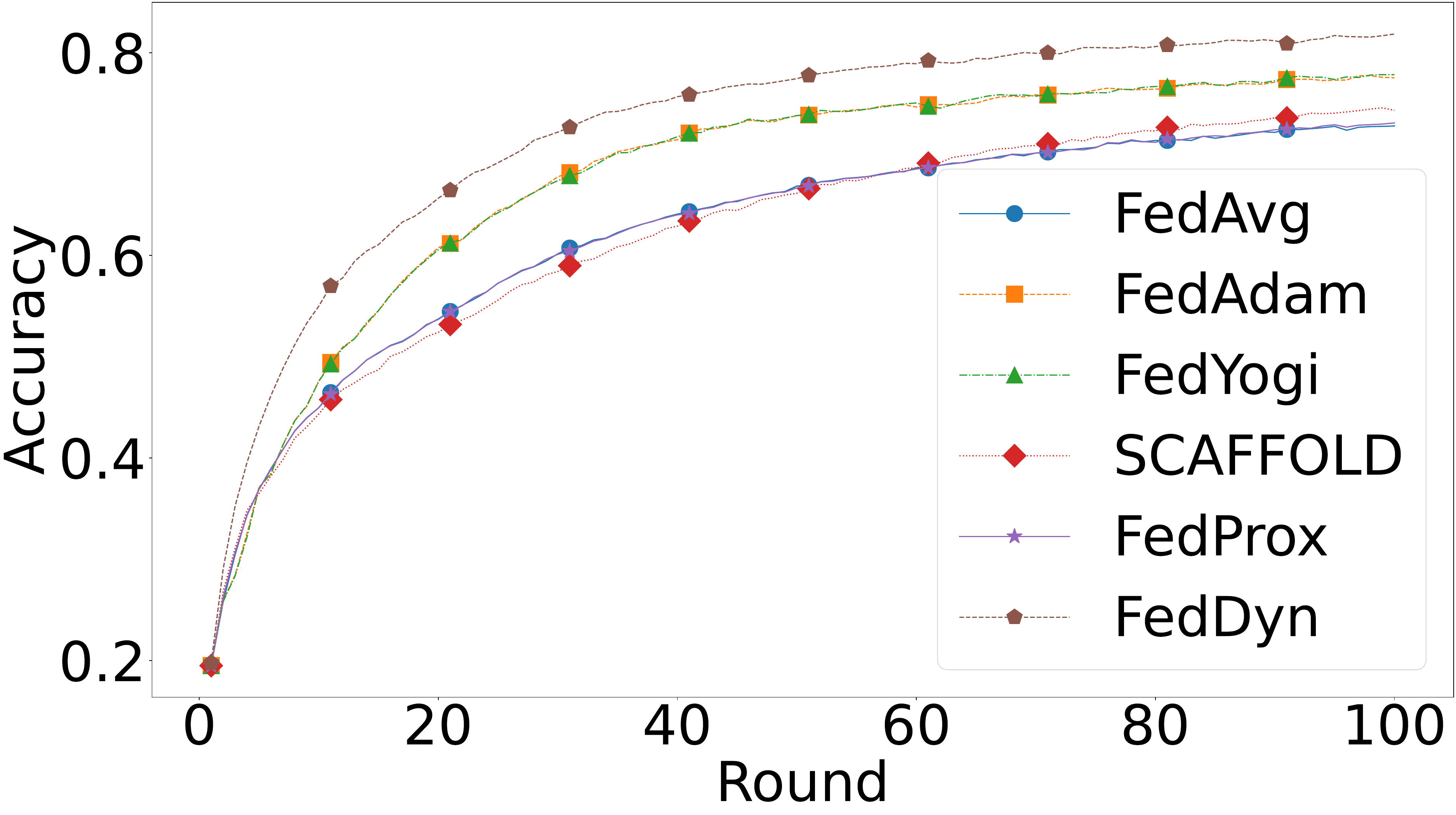}\label{fig:tst_acc_vs_round_cifar10_dir0.3}}
\hfill
\subfloat[Test accuracy vs. time]{\includegraphics[width=0.45\columnwidth]{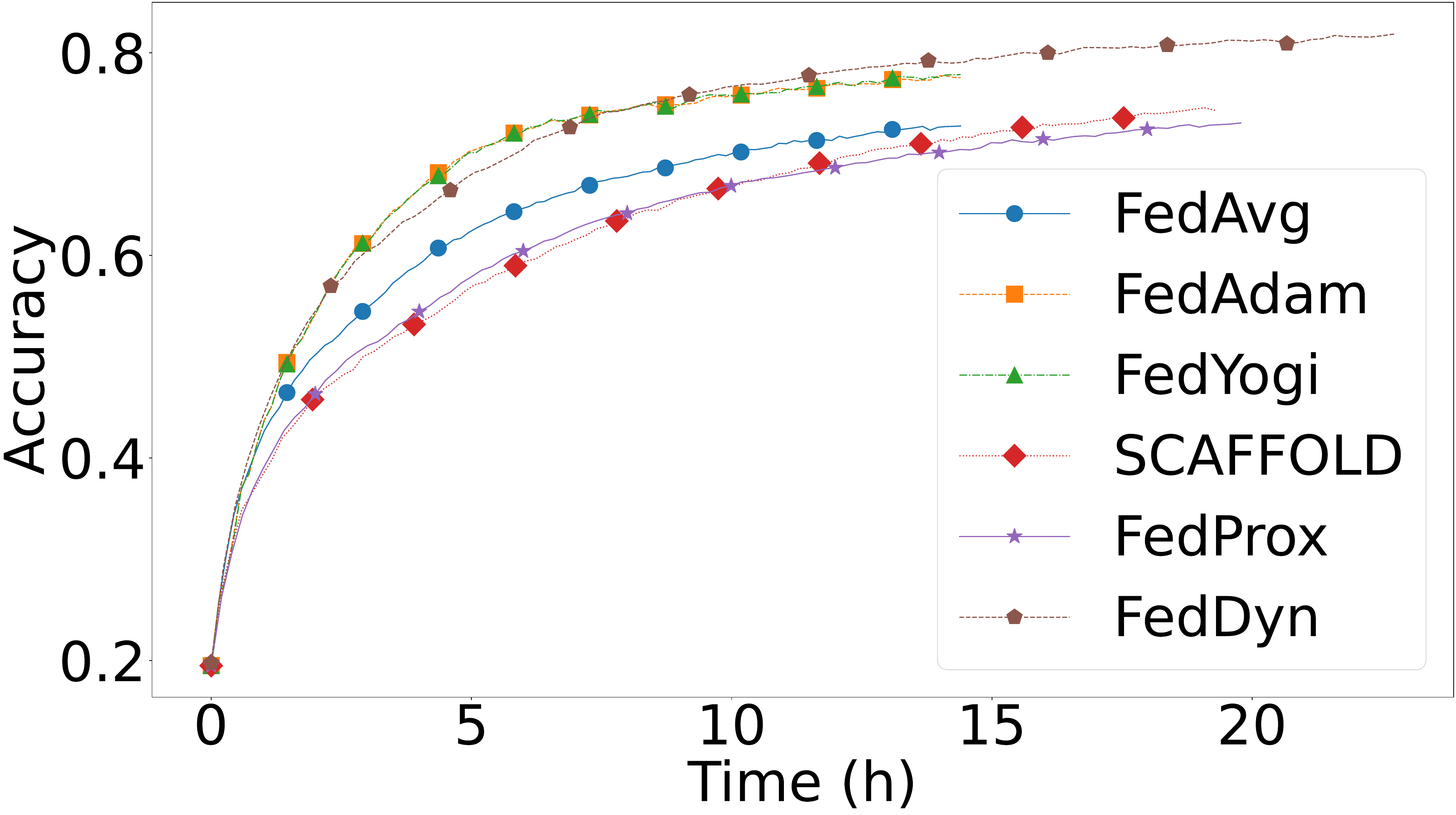}\label{fig:tst_acc_vs_time_cifar10_dir0.3}}
\caption{Test accuracy for CIFAR-10 dataset with \dirich{0.3} and 100 clients. We compare the accuracy across algorithms by choosing round or time on the x-axis.}
\label{fig:tst_acc_cifar10_dir0.3}
\end{figure}

\subsection{General Performance}
\label{subsec:general_performance}

To get a better idea of how the algorithms perform in terms of accuracy, we run them over 100 rounds while keeping track of the test accuracy of the global model at every round.
In particular, in order to understand how the algorithms work when resources are constrained, we run the experiment by using 100 clients on a machine with 32 CPUs and 256 GB of RAM. %
We make data distribution heterogeneous and non-IID by setting the parameter ($\alpha$) of Dirichlet distribution (denoted as $\mathrm{Dir}(\alpha)$) to 0.3 (i.e., $\alpha = 0.3$).

\takeaway{Different algorithms can exhibit significant difference in runtime to finish the same number of rounds.}

As shown in \fref{fig:tst_acc_vs_time_cifar10_dir0.3}, SCAFFOLD, FedProx, and FedDyn take 34.0\%, 37.4\% and 57.9\% longer to complete the training than FedAvg, respectively.
These results clearly indicate that the algorithms have different computational complexity and potentially pose important implications on time-to-accuracy under different hardware settings. We explore this aspect in \secref{s:runtime}.

\takeaway{Accuracy-to-round can be a misleading metric in evaluating the performance of an FL algorithm.}

In the federated learning community, accuracy-to-round (accuracy obtained after a certain number of rounds) is often adopted as a key performance metric
because higher accuracy-to-round can generally mean smaller number of communication rounds and hence lower communication overheads.
However, this metric ignores computation overhead in each round. This metric therefore fails to capture the fundamental trade-off between communication and computation overheads.
By comparing \fref{fig:tst_acc_vs_round_cifar10_dir0.3} and \fref{fig:tst_acc_vs_time_cifar10_dir0.3}, we observe the limitation of the metric. While FedDyn clearly stands out in terms of accuracy-to-round (\fref{fig:tst_acc_vs_round_cifar10_dir0.3}), the time-to-accuracy of FedYogi and FedAdam is comparable to FedDyn's (\fref{fig:tst_acc_vs_time_cifar10_dir0.3}) until their training completes (around the 14.38 hour mark). %
FedYogi and FedAdam are server-side optimization algorithms, and they do not require modification in the loss function, so they take less time to complete a round at client than FedDyn.
This essentially implies that they can learn for more 
rounds and further improve the accuracy before FedDyn finishes training.

\subsection{Execution Runtime}
\label{s:runtime}

As runtimes can be different across algorithms (discussed in \secref{subsec:general_performance}), it becomes important to understand how the runtime is affected under different hardware.
We look into this further by focusing on the client-side training runtime, since it drives the total runtime of a round.
Therefore, we run experiments by using one client and one server under four different kinds of compute resources (CPU and NVIDIA's T4, V100, A100).
As mentioned in \secref{sec:preliminaries}, some algorithms (e.g., FedProx, SCAFFOLD and FedDyn) introduce more terms in their loss function, compared to the basic FedAvg algorithm.
For GPU experiments, only the client process is allowed to use one GPU.
For the CPU experiment, the client process is configured to only use one CPU.

\begin{figure}[t]
\centering
\subfloat[A100]{\includegraphics[width=0.24\columnwidth]{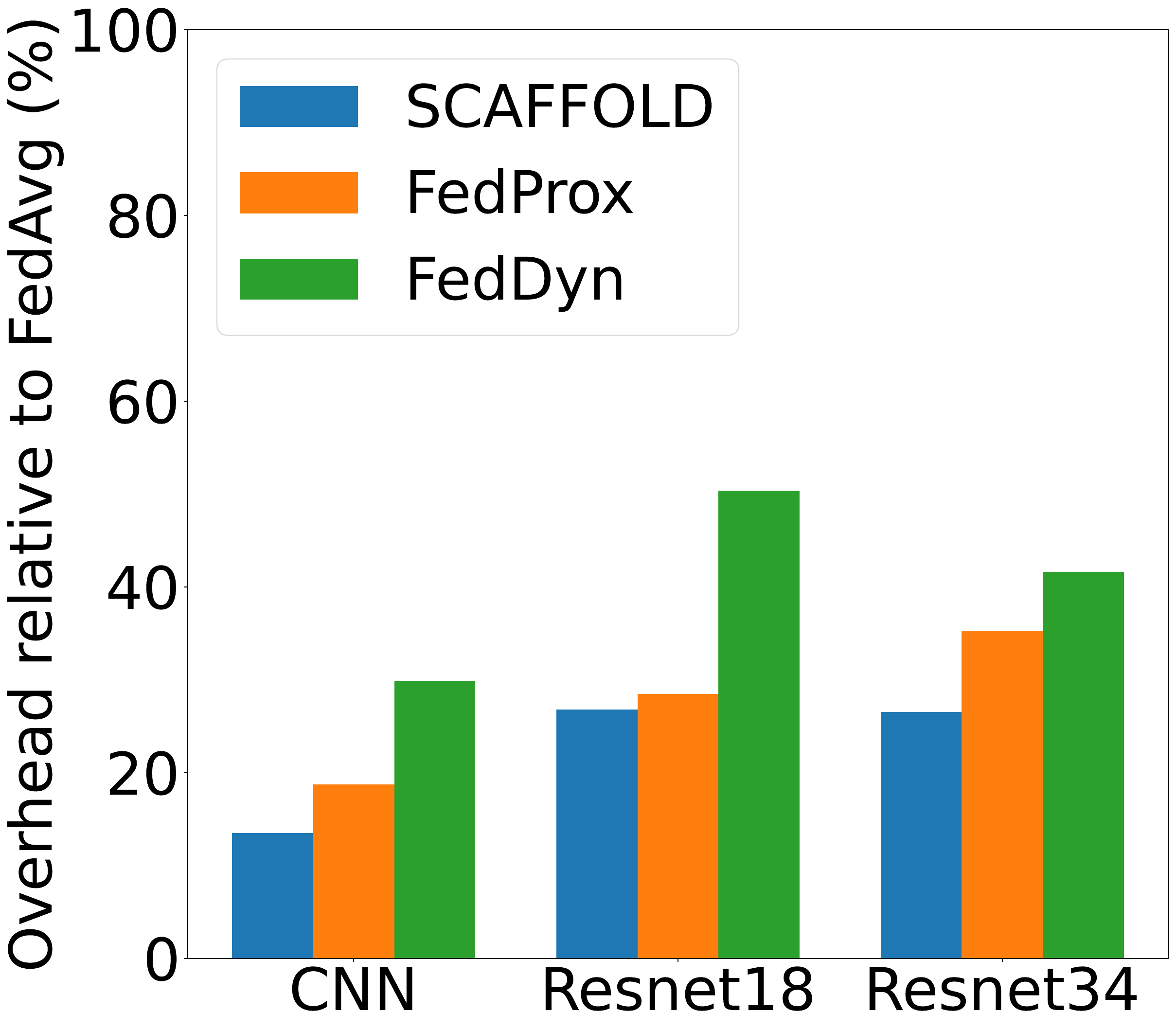}\label{fig:cifar10_runtimes_a100}}
\hfill
\subfloat[V100]{\includegraphics[width=0.24\columnwidth]{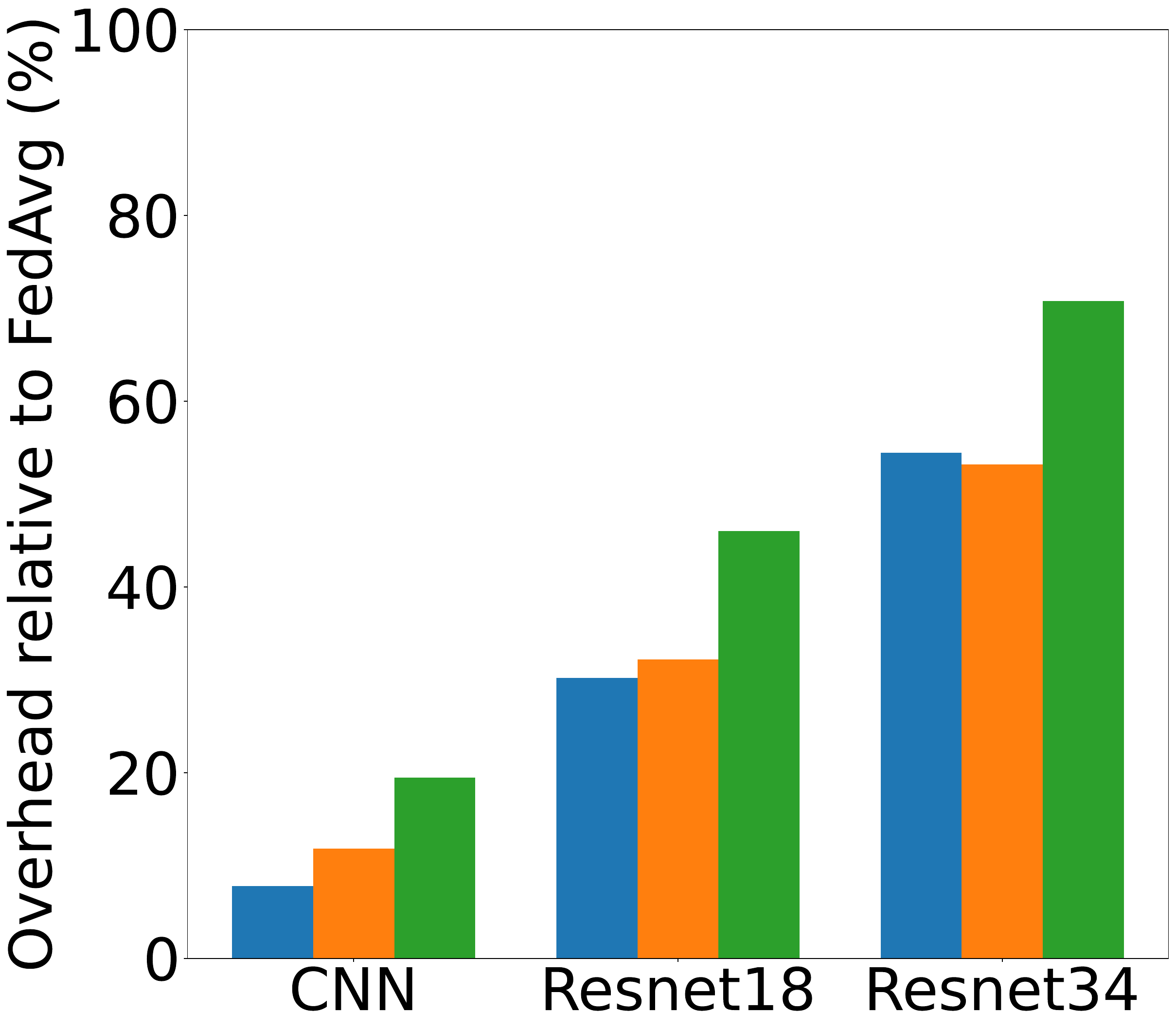}\label{fig:cifar10_runtimes_v100}}
\hfill
\subfloat[T4]{\includegraphics[width=0.24\columnwidth]{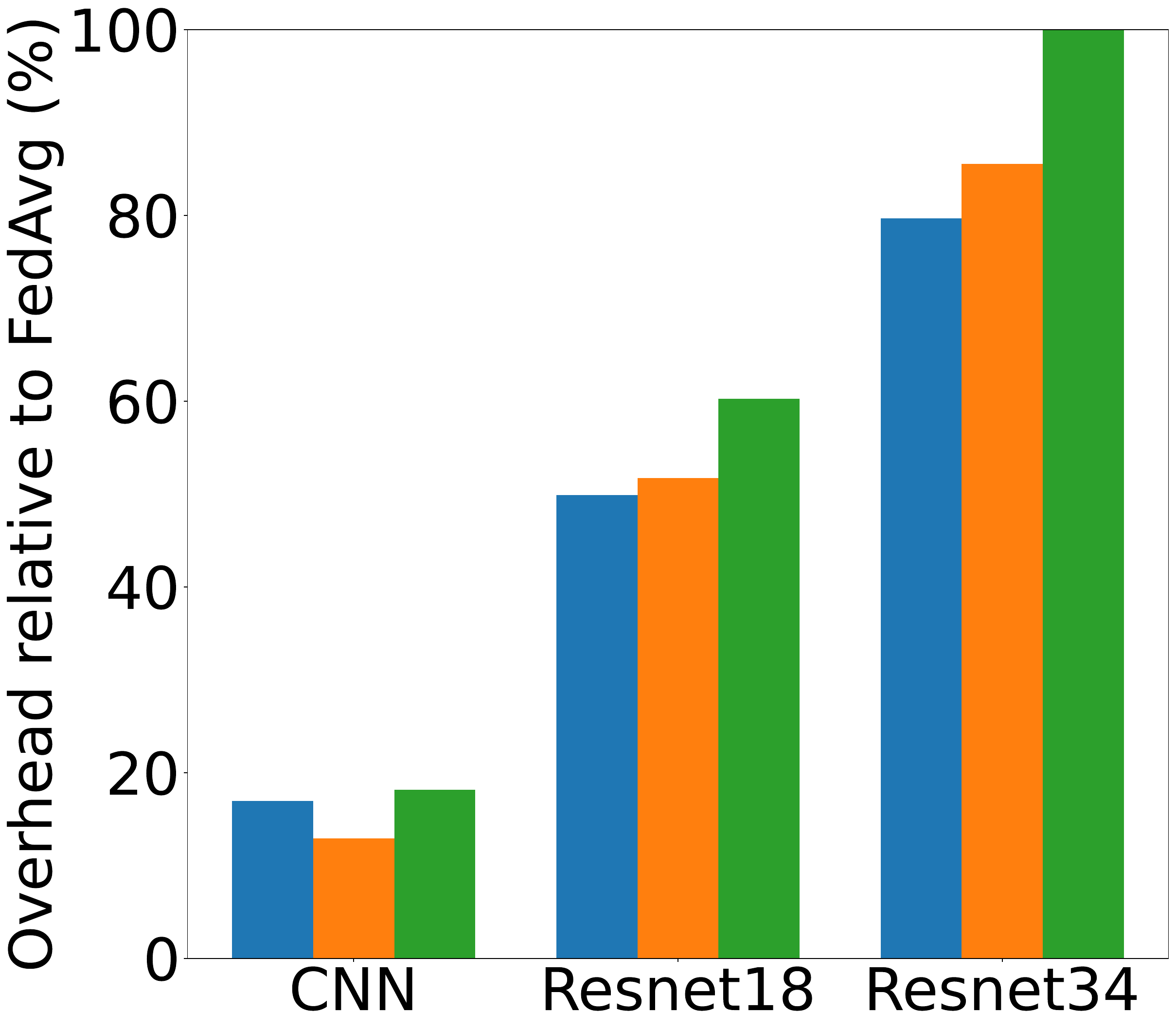}\label{fig:cifar10_runtimes_t4}}
\hfill
\subfloat[CPU]{\includegraphics[width=0.24\columnwidth]{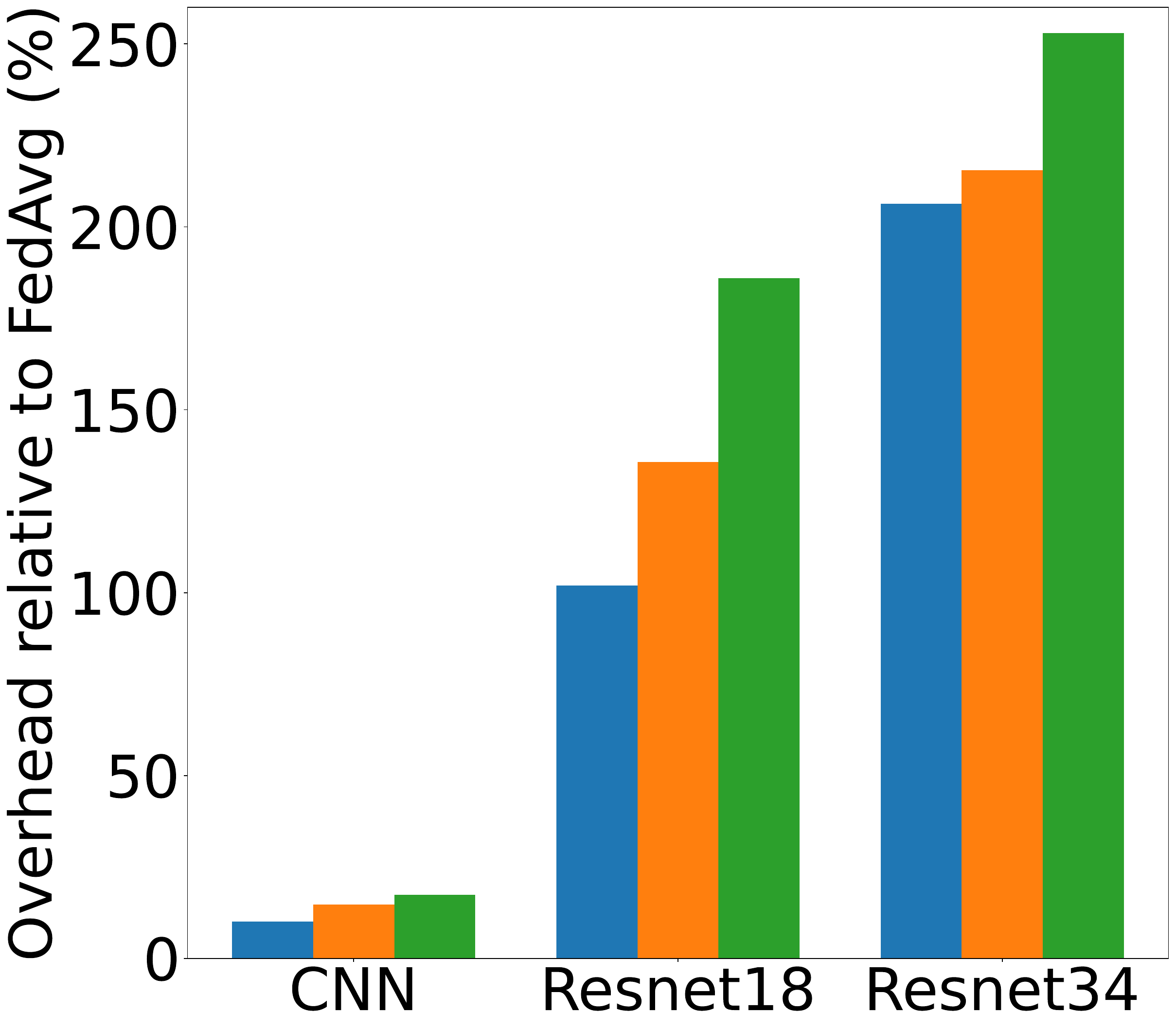}\label{fig:cifar10_runtimes_cpu}}
\caption{The relative runtime overhead of algorithms compared to FedAvg's runtime with CNN (798K parameters), ResNet18 (11.7M), and ResNet34 (21.8M). %
}
\label{fig:cifar10_runtimes}
\end{figure}
\begin{figure}[t]
\centering
\subfloat[A100]{\includegraphics[width=0.24\columnwidth]{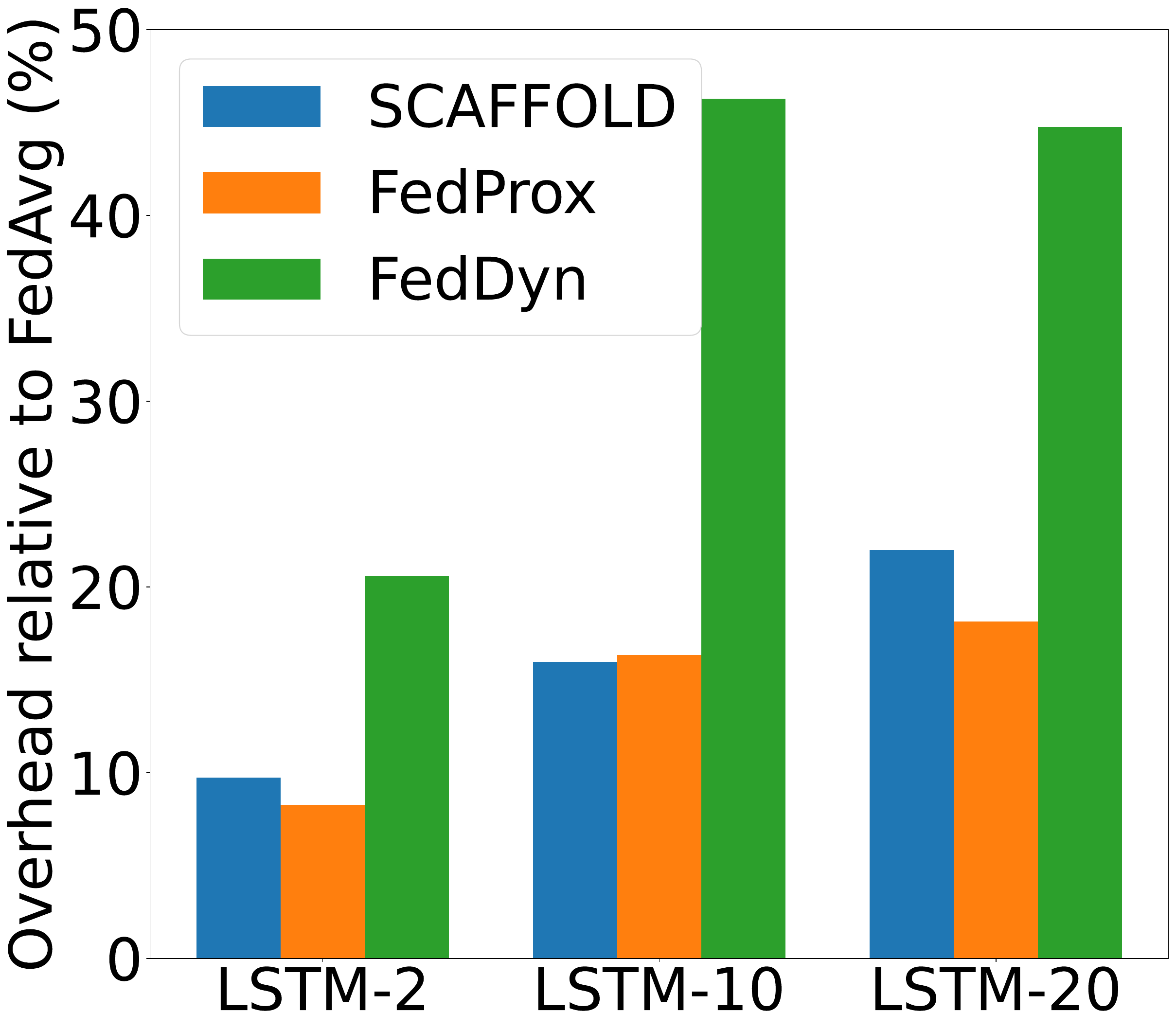}\label{fig:shakes_runtimes_a100}}
\hfill
\subfloat[V100]{\includegraphics[width=0.24\columnwidth]{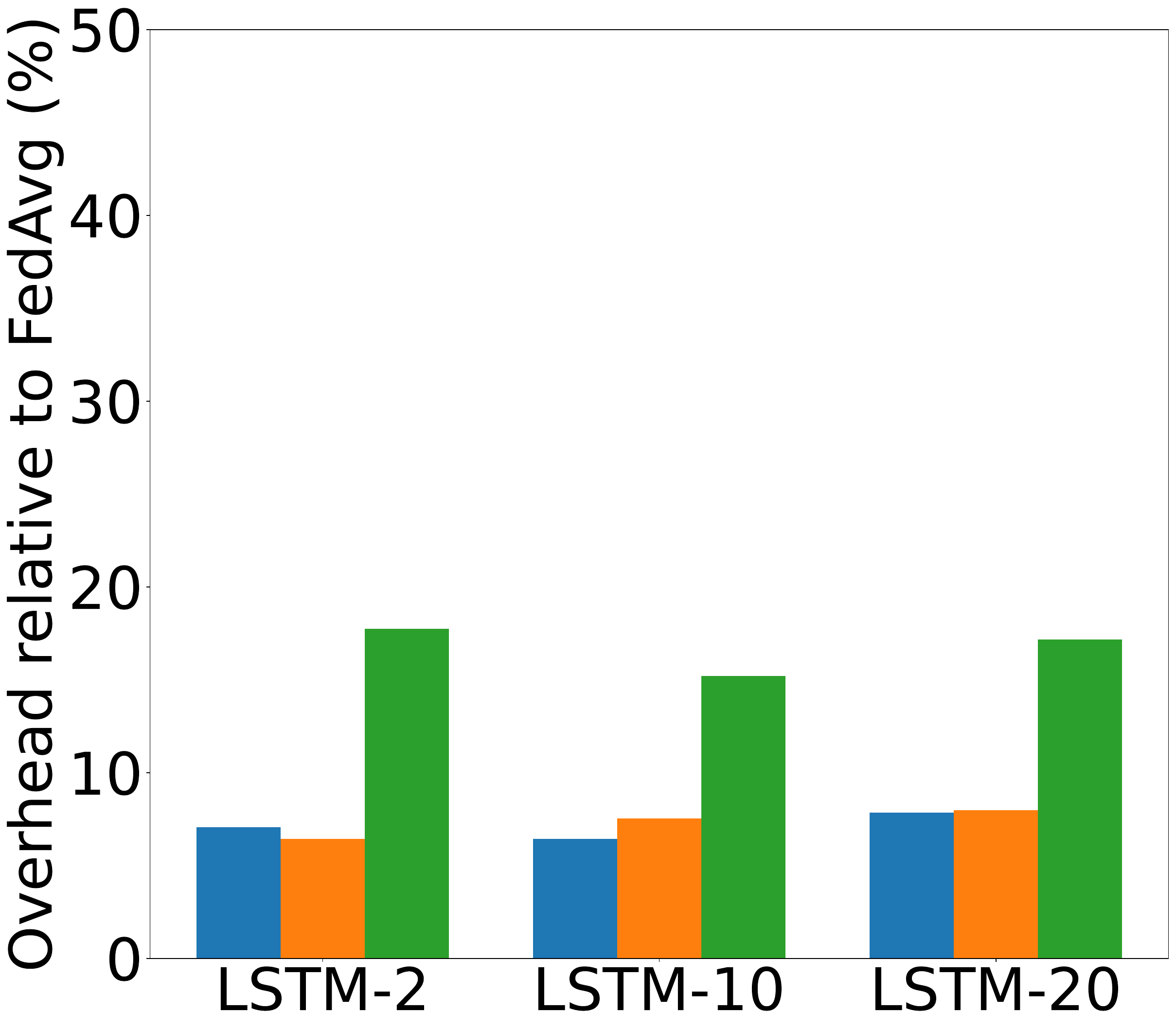}\label{fig:shakes_runtimes_v100}}
\hfill
\subfloat[T4]{\includegraphics[width=0.24\columnwidth]{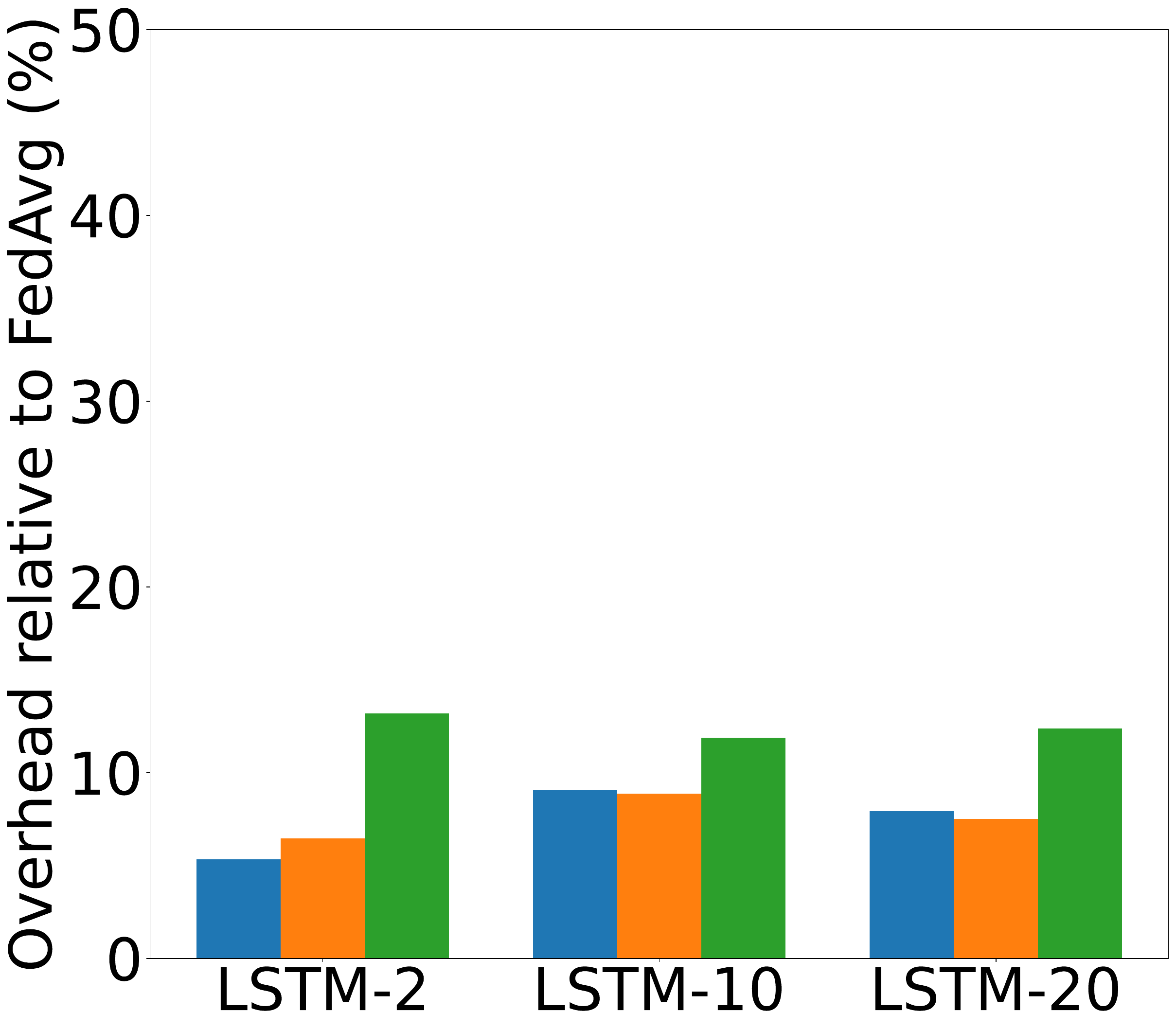}\label{fig:shakes_runtimes_t4}}
\hfill
\subfloat[CPU]{\includegraphics[width=0.24\columnwidth]{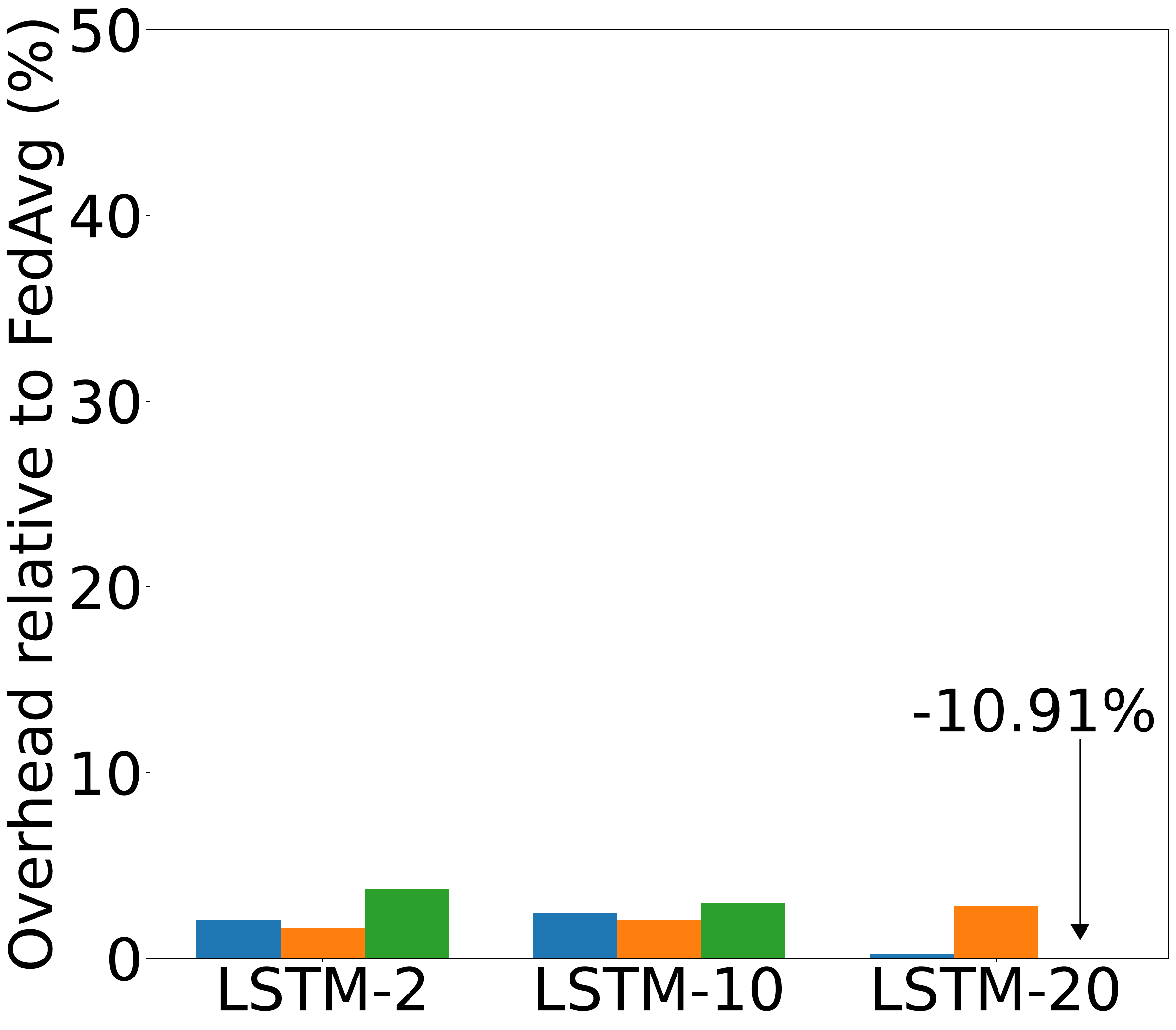}\label{fig:shakes_runtimes_cpu}}
\caption{The relative runtime overhead of algorithms compared to FedAvg's runtime with LSTM-2 (134K parameters), LSTM-10 (780K), and LSTM-20 (1.59M).
}
\label{fig:shakes_runtimes}
\end{figure}

\takeaway{FedDyn is the most computation-heavy algorithm among the algorithms under consideration.}

Figures~\ref{fig:cifar10_runtimes} and \ref{fig:shakes_runtimes} show the computation overhead of the algorithms running on the CIFAR-10 and Shakespeare datasets,
respectively. In the figures, the overhead is represented in a relative form (i.e., a relative overhead compared to FedAvg). The absolute runtime increases as the hardware becomes less powerful. As shown in the figures, FedDyn is the most expensive algorithm across most of the hardware types and model architectures. For instance, when run on a single CPU using ResNet34, the increases in runtime are 206.26\%, 215.41\%, and 252.90\% for SCAFFOLD, FedProx, and FedDyn, respectively. This demonstrates that the high overhead of FedDyn stems from the more complex regularizer than other algorithms for the loss function in the algorithm (as described in \eqref{feddyn_loss}).

\takeaway{Algorithm overheads are influenced by both hardware and neural network architecture.}

We investigate how the hardware and neural architecture influence the runtime of the algorithms with three convolutional neural networks (CNN, ResNet18 and ResNet34) and three recurrent neural networks (LSTM-2, LSTM-10, and LSTM-20).
As shown in \fref{fig:cifar10_runtimes_a100}, when the algorithms are tested on more powerful resources (e.g., A100), the computation overhead is lower: the maximum runtime difference with respect to FedAvg is about 50\%. As the computing power decreases (from \fref{fig:cifar10_runtimes_v100} to \fref{fig:cifar10_runtimes_cpu}), the runtime gap becomes larger: the maximum runtime difference is about 252.9\% in case of FedDyn with ResNet34. This suggests that the basic FedAvg is more suitable than other algorithms for less powerful devices.

However, \fref{fig:shakes_runtimes} presents the opposite trend. As the computing power decreases, the runtime difference among the algorithms also diminshes. We observe the largest runtime overhead under the A100 setting (\fref{fig:shakes_runtimes_a100}) while the maximum runtime overhead is merely 4\% under the CPU setting. Even with LSTM-20, FedDyn's runtime is 10.91\% faster than FedAvg; we also run the same experiment across three different machines and obtain around 4.1\% execution runtime speedup compared to FedAvg, thereby validating this observation.
Under less powerful devices, the small difference in overhead may be attributed to the fact that the parameters in an LSTM are reused frequently in the neural network, which therefore makes the complexities of the FL algorithms negligible for the runtime. On the other hand, since parameters are limited to being used in one layer for other neural networks such as CNN and ResNet, the limited computing power aggravates the runtime of the expensive algorithms such as FedDyn. Therefore, if a recurrent neural network is employed in FL training, the algorithms with the highest complexity (e.g., FedDyn) can still be a viable option for less powerful devices.

\subsection{Communication Overheads}

Communication overhead is an important factor when choosing an algorithm.
We capture the communication overhead across the algorithms by running an experiment with one client and one server.
We use a CNN architecture (798K parameters) for the CIFAR-10 dataset and the LSTM-2 architecture (134K parameters) for the Shakespeare dataset.
We only report the results of FedAvg and SCAFFOLD as other algorithms have the same communication overhead as FedAvg in terms of the volume of data exchanged between the server and client.

\takeaway{The communication overhead of SCAFFOLD is two times as large per round as that of all other algorithms considered.}

\tref{tbl:communication} shows the exact amounts of data exchanged by the algorithms we consider. 
The 100\% increase in communication between FedAvg and SCAFFOLD is due to the control variates used during client-side training in SCAFFOLD.
These control variates need to be shared over the communication network. Since the variates are added to the gradient at every client-side training step, they have the same size as the model weights. Therefore SCAFFOLD's communication overhead becomes the highest among the algorithms.

\begin{table*}[t]
\centering
\begin{tabular}{@{}lccr@{ }rr@{ }r@{}}
\toprule
\multirow{2}{*}{Architecture} & Number of & & \multicolumn{4}{c}{Algorithm} \\
\cmidrule{4-7}
    & Parameters & & \multicolumn{2}{c}{FedAvg (baseline)} & \multicolumn{2}{c}{SCAFFOLD} \\
\midrule
\multirow{2}{*}{LSTM-2} & \multirow{2}{*}{134K} & Sent & 25.89 MB & (0.00\%) & 51.78 MB & (99.97\%) \\
 & & Received & 25.89 MB & (0.00\%) & 51.78 MB & (99.98\%) \\
\midrule
\multirow{2}{*}{CNN} & \multirow{2}{*}{798K} & Sent & 153.87 MB & (0.00\%) & 307.74 MB & (100.00\%) \\
 & & Received & 153.87 MB & (0.00\%) & 307.74 MB & (100.00\%) \\
\bottomrule
\end{tabular}
\caption{Average megabytes communicated between the server and one client over 100 rounds for different algorithms datasets. We use the Shakespeare dataset for the LSTM-2 architecture, and the CIFAR-10 dataset for the CNN architecture.}
\label{tbl:communication}
\end{table*}

\begin{figure}
\centering
\subfloat[CIFAR-10; left: non-IID, right: IID]{\includegraphics[width=0.45\columnwidth]{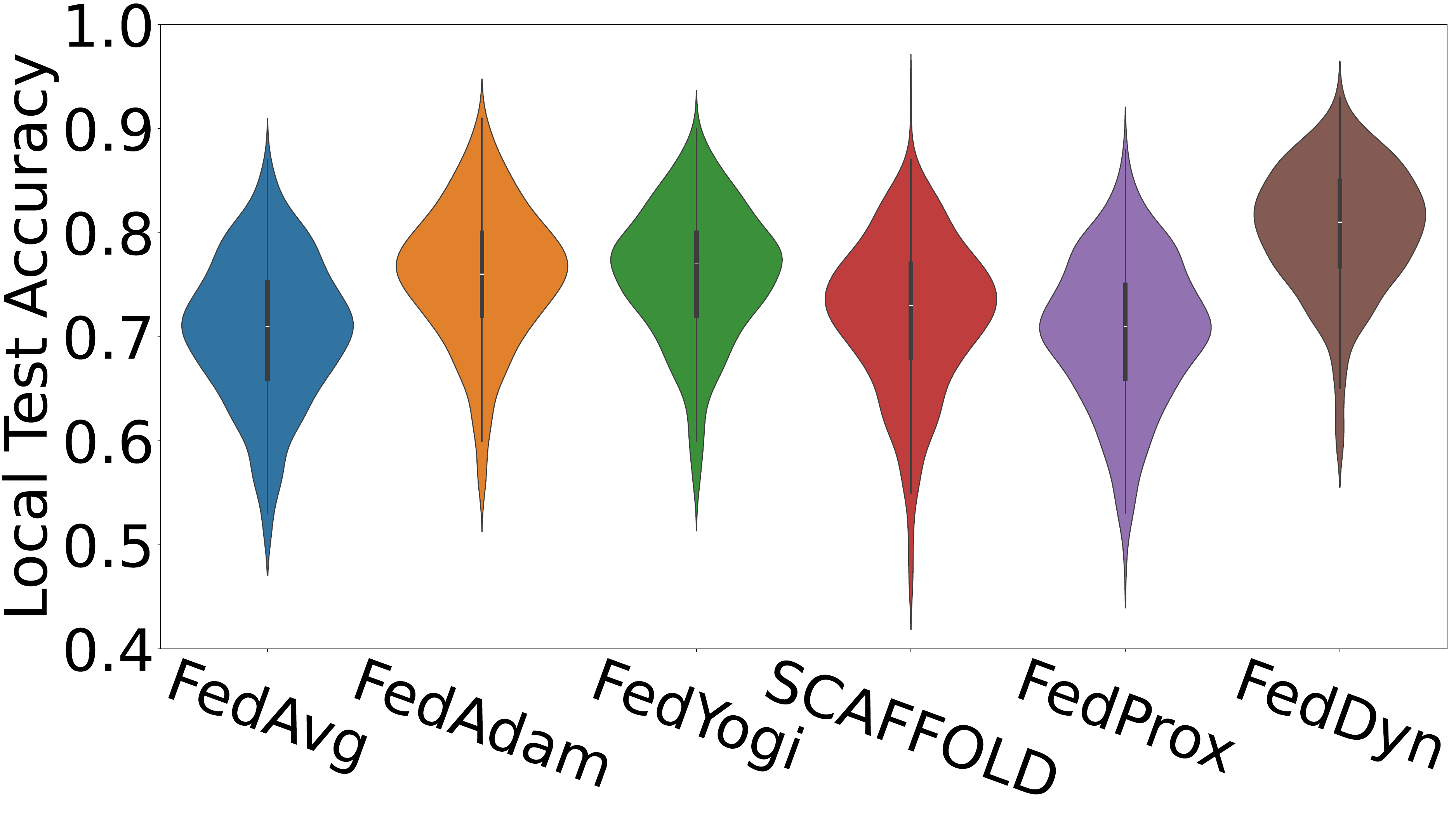}
\hspace{.2in}
\includegraphics[width=0.45\columnwidth]{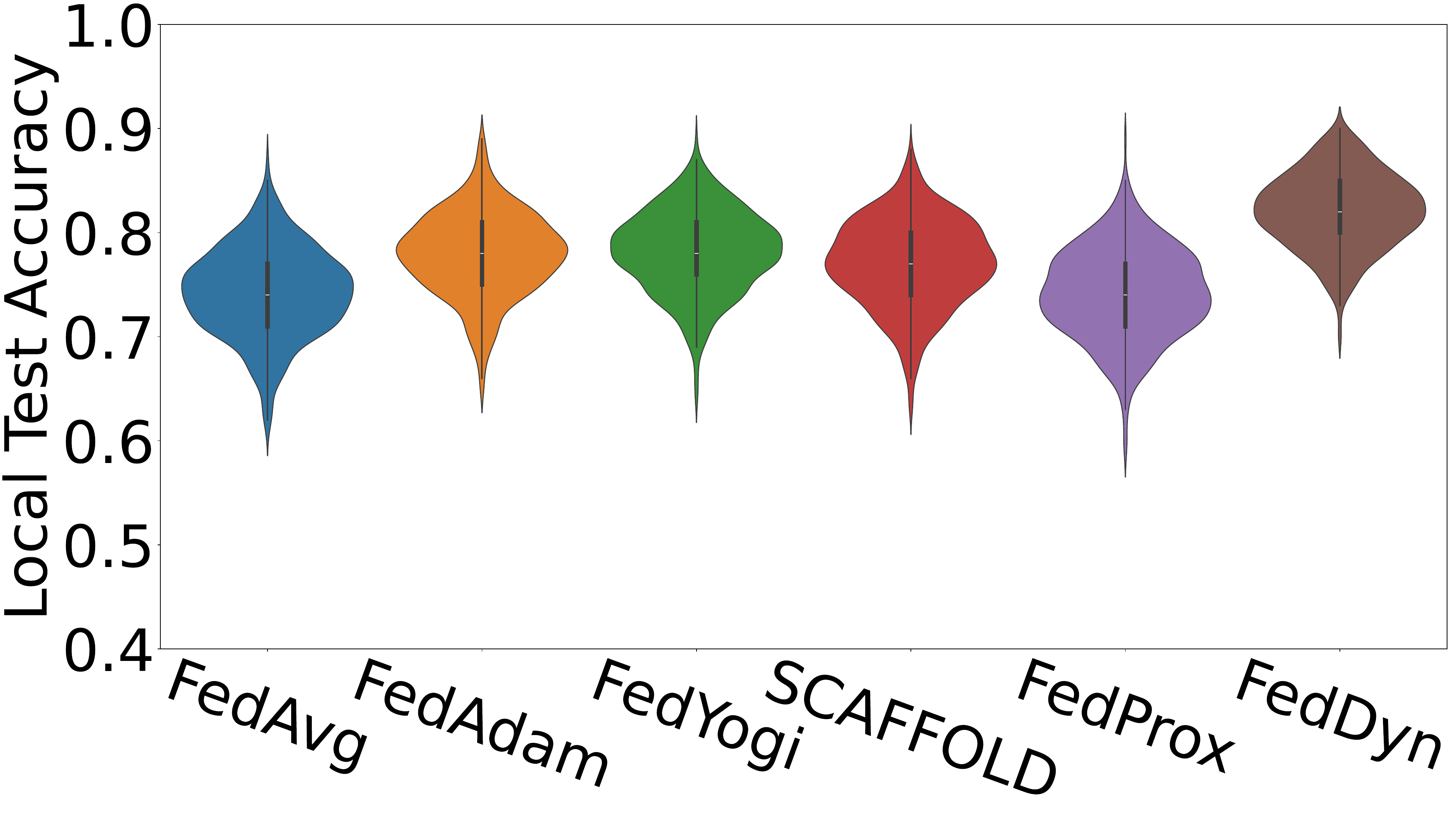}\label{fig:cifar10_psac}}
\hfill
\subfloat[Shakespeare; left: non-IID, right: IID]{\includegraphics[width=0.45\columnwidth]{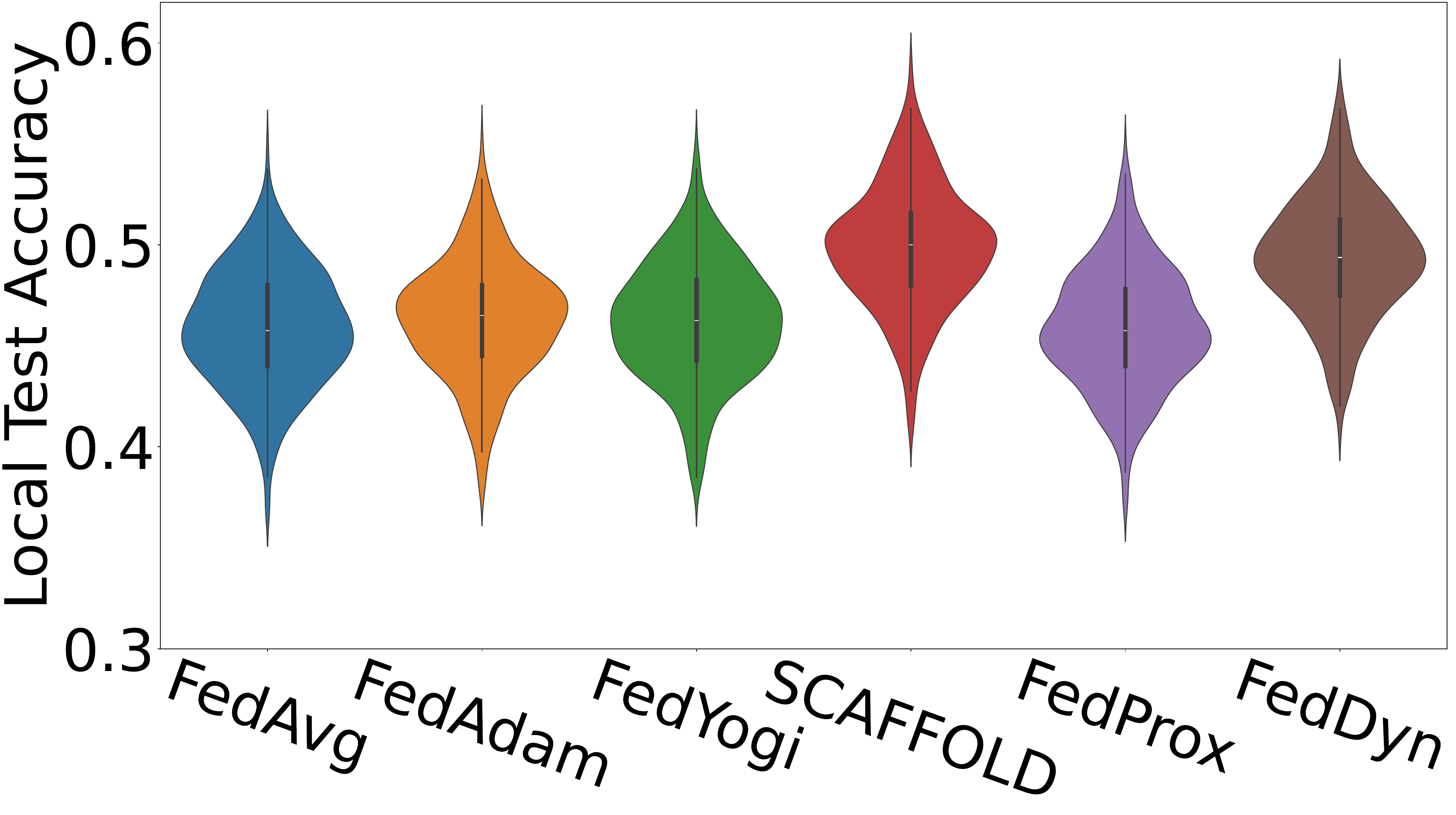}
\hspace{.2in}
\includegraphics[width=0.45\columnwidth]{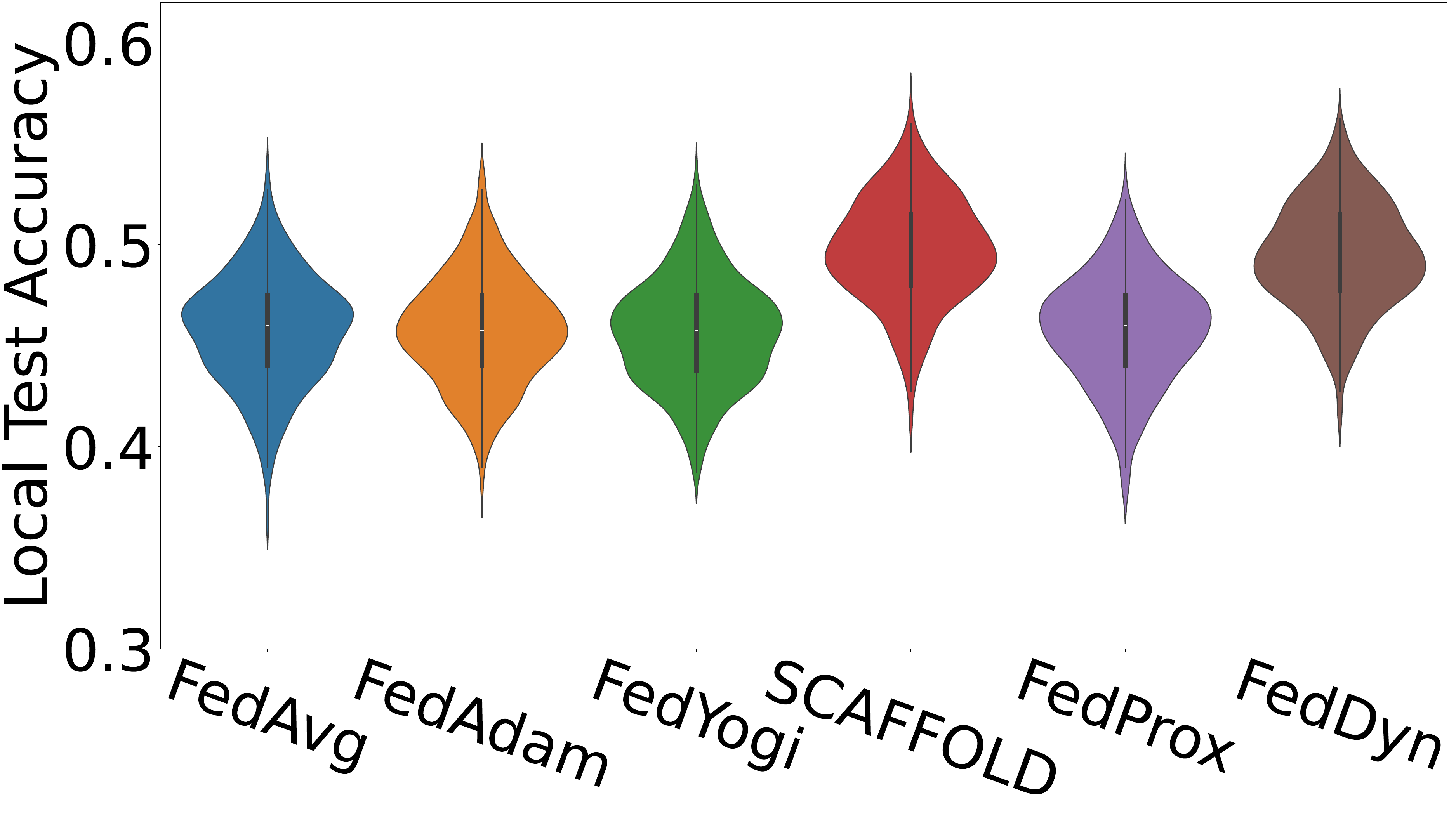}\label{fig:shakespeare_psac}}
\caption{Violin plots of local test accuracies for CIFAR-10 and Shakespeare for 5 different trials (a total of 500 values for each violin plot). These plots represent the distribution of local test accuracies across different runs.}  %
\label{fig:psac}
\end{figure}

\subsection{Performance Stability Across Clients}
\label{s:stability}

A typical process to evaluate a global model is to apply it with an isolated test dataset. This offers a simple view on the performance of the model. However, when the model is deployed in reality, it may face a diverse set of data, thus making its performance fluctuate across deployment settings (e.g., devices and users).
The less performance variation means that the model provides a consistent performance across the deployment settings, which is a desirable characteristic of an ML model. To quantify this characteristic, we define performance stability as the standard deviation of the accuracy values across clients.  A lower standard deviation would mean the global model is fairer across local test sets. Given the variety of FL algorithms, we investigate the performance stability of the algorithms of interest.

We generate IID and non-IID data distributions for 100 clients with the CIFAR-10 and Shakespeare datasets. For the CIFAR-10 dataset, we use \dirich{0.3} for the non-IID distribution, and generate balanced class distributions for IID. We use the Shakespeare dataset's natural non-IID distribution (a device is a character), and generate the IID version by pooling samples from all characters, as in \cite{feddyn}. %
We keep the global test set separately and only use it to confirm that training led to similar results as with the complete dataset.
Then, for each client we perform a random 80-20 local train-test split.
We run the training process by only using the train part of the train-test split to 100 rounds, and then test the final global model on each of the local test sets.
We measure the performance stability by averaging the standard deviations across 5 trials.
In \fref{fig:psac}, we include violin plots of all 500 measured local test accuracies.
The figure not only provides an idea of the distribution of local test accuracies, but also depicts the performance stability across the algorithms. The lower accuracy standard deviation means better performance stability.

\takeaway{FedDyn achieves the best performance stability among all the algorithms.}

FedDyn obtains 5-14.7\% lower standard deviation of accuracy than FedAvg's. Other algorithms excluding FedAvg perform 1-9\% worse than FedDyn. While FedDyn, FedProx, and SCAFFOLD all have client-side regularization terms, FedDyn is the only algorithm that performs better than FedAvg across all cases. 

\takeaway{SCAFFOLD is more prone to class imbalances in terms of performance stability.}

Our experiment results in \fref{fig:psac} show that SCAFFOLD's standard deviation is 2.5-4.3\% higher than FedAvg's across all non-IID distributions from the CIFAR-10 and Shakespeare datasets. SCAFFOLD only beats FedAvg in case of the IID distribution, which suggests that it does not handle class imbalances effectively across clients.

\takeaway{Server-side optimization algorithms can be a good alternative to client-side optimization algorithms.}

Server-side optimization algorithms such as FedAdam and FedYogi obtain better performance stability than FedAvg across all the distributions and datasets. Because SCAFFOLD and FedProx often fail to achieve better stability than FedAvg, this implies that the server-side algorithms are more stable and cost-efficient than SCAFFOLD and FedProx.

\takeaway{Performance stability across clients improves as the distributions become more IID.}

Performance stability turns out to be more consistent across clients as the test set distributions are more similar to each other in IID settings. For instance, for CIFAR-10, SCAFFOLD's accuracy standard deviation of the IID distribution decreases by almost 41.8\%, compared to that of the non-IID distribution. Other algorithms also see the similar reduction in the accuracy standard deviation (40.3\% for FedDyn and 38.6\% for FedAvg).

\begin{table}[t]
\centering
\begin{tabular}{@{}lcccccccc@{}}
\toprule
Algorithms & IID & \dirich{10} & \dirich{5} & \dirich{1} & \dirich{0.6} & \dirich{0.5} & \dirich{0.4} & \dirich{0.3} \\
\midrule
SCAFFOLD & 10\% & 60\% & 80\% & 100\% & 100\% & 100\% & 100\% & 100\% \\
FedDyn & 0\% & 0\% & 0\% & 50\% & 80\% & 100\% & 80\% & 40\% \\
\bottomrule
\end{tabular}
\vspace{0.1cm}
\caption{Catastrophic failure rates in the absence of gradient clipping. For each distribution generated from the CIFAR-10 dataset, the catastrophic failure rate is recorded out of 10 runs. All other algorithms face no failure across data distributions while FedAdam experiences 20\% failure rate only in the case of the \dirich{0.4} distribution.
}
\label{tbl:grad_clip_two}
\end{table}

\subsection{Impact of Gradient Clipping}

A catastrophic training failure occurs when the accuracy decreases by more than half in a single round~\cite{cohort}. 
The catastrophic failure hinges on a variety of factors including learning rate, distribution, and neural network architecture.
Gradient clipping is used to make federated learning algorithms more stable and avoid catastrophic failures.
However, the technique is not for all types of models. It is in general useful for the cases where models' weights can explode during the training. Models such as RNN, LSTM and transformers can present such a risk~\cite{grad_clip_acc_train}.
As it introduces an additional hyperparameter (the maximum norm of a gradient step),
a failure to setting a proper value for this parameter may lead to little learning and inefficient resource usage.
Therefore, in this experiment, our goal is to understand the stability of algorithms when gradient clipping is disabled from the client-side training process. We also vary data heterogeneity by employing Dirichlet distributions
to test the effect of data heterogeneity among clients for catastrophic failure rate.

\takeaway{Complex client-side optimizers experience catastrophic failure with high probability.}

Most algorithms we test (FedAvg, FedYogi, and FedProx) experience no failure. FedAdam has no failure except for one case with the \dirich{0.4} distribution; it has a failure rate of 20\%. SCAFFOLD and FedDyn are the only algorithms that exhibit higher failure rates. \tref{tbl:grad_clip_two} shows that both algorithms experience 40\% or higher failure rates when $\alpha \le 1$ for the Dirichlet distribution.
These two FL methods are the only ones that preserve a state aside from the model, which increases training complexity. %
FedDyn maintains the term $\nabla L(\theta_k^{t-1})$ locally, and also maintains a history term $h_t$ on the server side which is used in aggregation.
SCAFFOLD has a control variate which is used in the local loss function.

\takeaway{Failure rates improve for IID distributions.}

From \tref{tbl:grad_clip_two}, we observe a failure rate of 100\% for SCAFFOLD in the distributions with 0.6--0.3 Dirichlet values. %
As the distributions, became more IID %
the failure rate monotonically declines.
FedDyn's failure rate is highest (100\%) at \dirich{0.5}, and declines on either side.
In both algorithms, the lowest failure rates (10\% for SCAFFOLD and 0\% for FedDyn) are obtained when the dataset distribution is IID. This meets our expectation since the IID case would lead the data to be similar across clients, which, in turn, would likely lead to smaller training updates. Therefore, the likelihood of overflow is smaller than that of the non-IID case. %

\takeaway{Failure rates improve with a smaller learning rate.}

Previously, there have been examples showing some correlation between gradient norm and learning rate~\cite{grad_clip_acc_train}, suggesting that a large learning rate could accelerate the growth in magnitude of the model weights.
Therefore, we also tested SCAFFOLD and FedDyn with a smaller learning rate.
In addition to our original learning rate $\eta=0.1$, 
we employ $\eta=10^{-\frac{3}{2}} (\sim 0.032)$ since the value is used in \cite{fedopt}.
With this smaller learning rate, the failure completely disappears for SCAFFOLD and FedDyn across all distributions. This indicates that the failure rate, in the absence of gradient clipping, is highly sensitive to the learning rate, since the two learning rates we tested only differ by a factor of $10^{-\frac{1}{2}}$.

\section{Discussion}
\label{sec:discussion}

\textbf{Choosing algorithm.}
This paper focuses on generic synchronous FL algorithms and excludes algorithms which try to optimize metrics such as fairness. Here we discuss what to consider beyond accruacy in choosing an FL algorithm.

The algorithms considered in this paper are frequently evaluated with respect to round.
There is a cost associated with modifying the client-side loss functions, which can result in the round being an unfair measure of work (see \fref{fig:tst_acc_vs_round_cifar10_dir0.3} and \fref{fig:tst_acc_vs_time_cifar10_dir0.3}).
In the presence of an additional loss function component, the graph used for back propagation grows in size linearly related to the number of trainable parameters in the neural network.
Therefore, we observe the training runtime increase from SCAFFOLD, FedProx and FedDyn.
In some cases, this means algorithms with simpler server-side optimization may be cost-efficient.
This can be taken into account when the cost of training in one round needs to be minimized.

Additionally, consistent performance across clients can be important for an application.
We observe differences across optimizers for this metric.
FedProx is among the worst performing for this test.
In general, FedDyn, FedAdam, and FedYogi usually perform better.
We also notice that SCAFFOLD performs worse on all non-IID distributions.

Catastrophic failure is another factor to consider in choosing an algorithm. In particular, we learn that gradient clipping can affect optimizers severely.
That is, algorithms that preserve additional state apart from the global model weights are more likely to experience catastrophic failures while training without gradient clipping. We demonstrate algorithm's vulnerability with gradient clipping. As far as gradient clipping is concerned, it is useful to keep it enabled. However, there may be other vulnerabilities due to unknown factors (e.g., dataset heterogeneity degree, straggling client, and so on).
Thus, one should carefully assess the tradeoff between failure possibility and performance gain.

\textbf{Different types of FL approaches.} There are numerous FL approaches beyond the algorithms we considered in this paper. Oort~\cite{oort} employs client overselection and client's utility to expedite the training process. FedBalancer~\cite{fedbalancer} systematically picks `useful' data samples to achieve a faster convergence rate. FedBuff~\cite{fedbuff} trains the model asynchronously by aggregating local models at any point in time during the training without per-round synchronization barrier. REFL~\cite{refl} selects clients to improve resource-efficiency. One future direction is to study how the algorithms would work with these approaches.

\section{Conclusion}
\label{sec:conclusion}

In this paper, we empirically study six well-known federated learning algorithms in terms of time-to-accuracy, computation and communication overhead, performance stability across clients, and training instability. Our experimental evaluation reveals that no single federated learning algorithm outperforms across all the evaluation metrics under consideration. One of the most recent algorithms, FedDyn in general achieves the highest accuracy given a fixed number of rounds. However, it tends to require higher amount of computation resources than other algorithms, thus taking longer than other algorithms to finish the fixed number of rounds. In addition, FedDyn is likely to face some instability issue more frequently than other algorithms during training; we test this hypothesis by disabling gradient clipping. While FedDyn achieves the smallest accuracy standard deviation across clients (hence, the highest performance stability), other client-side optimization algorithms such as FedProx and SCAFFOLD obtain lower performance stability than server-side algorithms such as FedYogi and FedAdam. In general, those server-side optimization algorithms can be good alternatives as they tend to perform reasonably well with little extra computation overhead and run without failures in the absence of gradient clipping. We hope that our results can assist FL algorithm selection task for FL practitioners and encourage the community to build best practices for FL algorithm evaluation.

\bibliographystyle{plain}
\bibliography{reference}

\end{document}